\documentclass[10pt, a4paper, conference, compsocconf]{IEEEtran}
\usepackage[utf8]{inputenc}
\usepackage{cite}
\usepackage[cmex10]{amsmath}
\usepackage{url}
\usepackage{todonotes}
\usepackage{graphicx}
\usepackage{bm}

\title{Fully Convolutional Neural Networks for Page Segmentation\\of Historical Document Images}
\author{\IEEEauthorblockN{Christoph Wick and Frank Puppe}
\IEEEauthorblockA{Chair of Informatics VI\\
Department of Informatics, University of Würzburg, Germany\\
Email: \{firstname.lastname\}@uni-wuerzburg.de}
}
\date{November 2017}

\begin{document}

\maketitle

\begin{abstract}
    We propose a high-performance fully convolutional neural network (FCN) for historical document segmentation that is designed to process a single page in one step.
    The advantage of this model beside its speed is its ability to directly learn from raw pixels instead of using preprocessing steps e. g. feature computation or superpixel generation.
    We show that this network yields better results than existing methods on different public data sets.
    For evaluation of this model we introduce a novel metric that is independent of ambiguous ground truth called Foreground Pixel Accuracy (FgPA).
    This pixel based measure only counts foreground pixels in the binarized page, any background pixel is omitted.
    The major advantage of this metric is, that it enables researchers to compare different segmentation methods on their ability to successfully segment text or pictures and not on their ability to learn and possibly overfit the peculiarities of an ambiguous hand-made ground truth segmentation.
\end{abstract}


\section{Introduction}

In the digitalization pipeline of historical books the segmentation of a page into different regions such as pictures and texts is a crucial step for all further processing including optical character recognition (OCR).
Errors in the text segmentation, e. g. cropped or forgotten characters, directly affect the outcome of the OCR that tries to translate written or printed text into digitized characters.
Especially the segmentation of historical documents is challenging, because these documents suffer from degradation, different layouts or writing styles, and often include ornaments or decoration.

A performant segmentation algorithm whose outcome shall be used for OCR must be capable of separating background, text, and possibly images.
As improvement it performs fine grained semantic distinctions to a text block, such as headline, running text, page number, or marginalia, which are common text types in historical documents.

The basic approach for a segmentation algorithm is to assign a label to each pixel of the page.
Pixel-by-pixel semantic classifications based on convolutional nerual nets (CNNs) have widely been used (see e. g. \cite{journals/pami/FarabetCNL13} or \cite{conf/icml/DonahueJVHZTD14}).
Novel approaches \cite{journals/pami/ShelhamerLD17} use fully convolutional networks (FCNs) to learn and infer the whole image in one step.
This method does not require additional pre- or post-processing steps, such as superpixels (see e. g. \cite{journals/pami/FarabetCNL13}, \cite{conf/das/chen/2016}, \cite{conf/icfhr/0011SLHLI16}, \cite{chen2017convolutional}).

Our fully convolutional neural net is an adaptation of the U-Net proposed by Ronneberger et al. \cite{journals/corr/RonnebergerFB15}.
It consists of several convolution, pooling and deconvolution operations, but in contrary to the U-Net it does not use skip connections.
The encoding and decoding structure ensures that only information that encodes large regions in the page is forwarded.
The main advantage of the proposed network architecture is the prediction of a full image in one step.
Hence, the network can utilize all available information (e. g. the position of marginalia) to classify each pixel.
Therefore, the network is not only able to master the comparatively easy task of distinguishing periphery, page and text but can also successfully perform a fine-grained semantic distinction between a large variety of very similar classes.

\begin{figure}[t]
\centering
\includegraphics[width=0.8\linewidth]{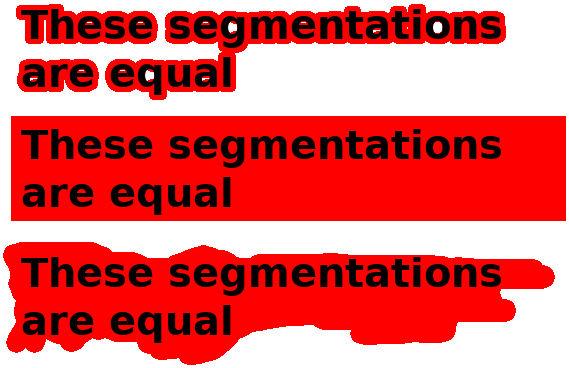}
\caption{
Different segmentations of the same text.
The first line segments almost each single character, the second line is segmented as a rectangle, and the third example is an arbitrary segmentation.}
\label{fig:equal_segmentations}
\end{figure}

For training and evaluating a segmentation algorithm, human annotated ground truth is required.
In those ground truth masks a region is usually labeled as a whole block, which contains fore- and background pixels (see middle of Figure \ref{fig:equal_segmentations} and column 3 of Figure \ref{fig:segmentation_results}).
For the purpose of OCR as step of the digitization pipeline, this segmentation is ambiguous.
All three segmentation examples in Figure \ref{fig:equal_segmentations} are equal in the sense that a further OCR algorithm gains all the required information for a successful recognition.
Obviously, the upper segmentation example loses information about the connected block, but it is rather simple for instance by a growing algorithm to join single lines or characters.
A standard Loss-Function as the goal of training is optimal if the network predicts exactly the provided ground truth, here the middle of Figure \ref{fig:equal_segmentations}.
The proposed learning algorithm ignores background labels and predicts only labels of the foreground.
For this purpose, all background labels are ignored in the loss function of the FCN.
Thus, the network is explicitly allowed to predict any label for a background pixel.

Obviously, the established metrics to evaluate such a model that are mostly variations on pixel accuracy and region intersection over union (IU) (see e. g. \cite{journals/pami/ShelhamerLD17} or \cite{chen2017convolutional}), can not be used.
If the model predicts the upper segmentation of Figure \ref{fig:equal_segmentations} but the ground truth is the middle of Figure \ref{fig:equal_segmentations} the score would be very low, although the segmentation would be perfect in the described sense of OCR post-processing.
For this purpose, we introduce a novel pixel based measure, that only tracks foreground pixels and ignores all background pixels.
This metrics predicts by construction the same value for all of the three example segmentations of \ref{fig:equal_segmentations}.
The lower bound is 0 if not a single pixels is labeled correctly and the maximum value is 1 if all foreground pixels are assigned with the ground truth.
To compare our FCN with other models we also evaluate the established pixel accuracy.

\section{Related work}
The field of document analysis of contemporary and historical documents is a challenging problem and has been addressed by many researchers.
Recently, algorithms based on CNNs that automatically learn features from the pixels of a page show superior success in this area compared to handcrafted features \cite{chen2017convolutional}, \cite{yang2017learning}.

FCNs are a new trend in general semantic segmentation and were successfully trained by \cite{conf/cvpr/LongSD15}, \cite{journals/corr/NohHH15}, or \cite{journals/pami/ShelhamerLD17}.
The decoding part of those networks usually consist of several layers of unpooling \cite{conf/iccv/ZeilerTF11} and deconvolutional (e. g. \cite{journals/corr/NohHH15}) operations.

Yang et al. \cite{yang2017learning} try to extract semantic structure from contemporary documents using a multimodal fully convolutional neural networks (MFCN).
They combine a simple FCN with several extensions, including a text embedding map that tries to input both the visual representation and the true text.
Moreover, they implement an additional unsupervised consistency task that forces the neural network to have similar activations in each individual rectangular region of the ground truth.

An adaptation of a FCN is the so-called U-Net introduced by Ronneberger et al. \cite{journals/corr/RonnebergerFB15}, which is applied to the field of biomedical image segmentation and outperforms existing methods for cell tracking by a large margin.
This network consists of several convolution and pooling layers, but only uses deconvolutions as upsampling operations instead of unpooling layers.
Furthermore, it uses skip connection from the encoder to the decoder part of the network to preserve high-level information for the decoder.

In the field of historical document analysis Chen and Seuret \cite{chen2017convolutional} proposed a three layer neural net with only one convolutional layer.
This network learns to predict the label of superpixels (see e.g. \cite{journals/pami/FarabetCNL13}) and outperforms methods that are based on Support Vector Machines  \cite{conf/icdar/ChenSLHI15}, \cite{conf/das/chen/2016} or Conditional Random Fields \cite{conf/icfhr/0011SLHLI16} and handcrafted features.

\section{Datasets}
For all our experiments we use several historical books (see Table \ref{tab:datasets}).
\emph{GW5060}, \emph{GW5064} and \emph{GW5066} are different editions of "The Ship of Fools", that were scanned during the Narragonien digital\footnote{\url{http://kallimachos.de/kallimachos/index.php/Narragonien}} project.
Although these data sets share the same content, they all have a different layout, font, images or size.
These books are already deskewed and cropped to a page without periphery.
The three further publicly\footnote{\url{http://www.fki.inf.unibe.ch/databases/iam-historical-document-database} and \url{http://diuf.unifr.ch/main/hisdoc/divadia}} available sets \emph{Parzival} \cite{conf/icdar/WuthrichLFIBVS09}, \emph{St. Gall} \cite{fischer2011transcription}, and \emph{G. Washington} \cite{vlavrenko2004holistic} consist only of a few annotated pages and are split in a fixed training size of 24, 21, and 10, respectively.
Those books are skewed and contain periphery that must be segmented, too. 
An exemplary page of G. Washington and GW5060 is shown in Figure \ref{fig:segmentation_results}.

The provided ground-truth segments a page into various regions: background, page, running text, marginal, page number, running head, chapter heading, motto, and signature mark.
Each single book uses only a subset of all available categories (see Table \ref{tab:datasets}).

\begin{table}[!t]
    \caption{Details of the used data sets. The ratio is given as $w/h$.}
    \label{tab:datasets}
    \centering
    \begin{tabular}{c|cccc}
    \hline
        Data set & Image size (pixels) & Ratio & Pages & Labels\\
    \hline
        GW5060 & $\approx 1900 \times 2980$ & $0.64$ & 158 & 6 \\
        GW5064 & $\approx 2030 \times 2940$ & $0.69$ & 351 & 6 \\
        GW5066 & $\approx 1000 \times 1566$ & $0.64$ & 234 & 6 \\
    \hline
        Parzival & $2000 \times 3008$ & $0.66$ & 37 & 5 \\
        St. Gall & $1664 \times 2496$ & $0.67$ & 60 & 5 \\
        G. Washington & $2200 \times 3400$ & $0.65$ & 20 & 5 \\
    \hline
    \end{tabular}
\end{table}

\section{The model}

\subsection{Pre-processing}
The training of the model requires the input image and the target segmentation that is ground truth.
Since not all pages in a single book have the same dimensions, we fit each single image in a white page with a fixed ratio of $2/3$.
Afterwards the pages are scaled down for the experiments to a size of $260 \times 390$ pixels, which reduces both the computation time and the network size.
Note, that the model is evaluated on the full resolution images by scaling the predicted mask to the original image size.
We apply ocropy's\footnote{\url{https://github.com/tmbdev/ocropy}} binarization tool on each input image to provide a binary image which is used as bit-mask to separate fore- and background, that is black and white, respectively.
The target mask is chosen to be 0 at true background, that is white color in the binarized images, and otherwise set to the desired labels.
As stated, the 0 label will be ignored during the training, which allows the network to predict any color at these pixel positions.

An example input, binarized and provided ground truth page is shown in columns 1 to 3 of Figure \ref{fig:segmentation_results}, respectively.

\subsection{FCN Architecture}

\begin{figure*}[!t]
\centering
\includegraphics[width=\linewidth]{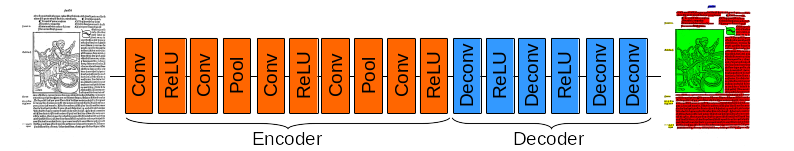}
\caption{Overview of the used FCN structure. The input image is a full page, the complete segmented image is produced as output in one step.}
\label{fig:net_structure}
\end{figure*}

Our proposed FCN network that is sketched in Figure \ref{fig:net_structure} consists of an encoder and a decoder structure based on convolution- and deconvolution-layers.
Deconvolution layers can basically be seen as an inverse convolution (see e. g. \cite{conf/cvpr/ZeilerKTF10}).
By setting the stride equal to the kernel size it can be used to increase the image resolution by that factor.
The proposed network consists of pooling layers with a kernel size and stride of 2, which is why the two middle deconvolutional layers also have kernel size and stride of 2 to increase the image dimensions in accordance.


The encoding consists of more convolutional layers than the decoder because the encoder has to process more granular information.
Each single convolutional layer has a kernel size of 5 and uses padding at the margin to keep the image resolution.
The number of filters in the convolution and deconvolution layers are 40, 60, 120, 160, 240 and 240, 120, 60, and 6, respectively.
The last deconvolution-layer is used to generate the prediction of up to six different labels.

The utilized cross entropy loss ignores all background labels, which is why the network is allowed to predict any label where the mask is background.

\subsection{Post-processing}
Since the output of the neural network is a mask that can predict arbitrary labels at the ignored background this outcome must be multiplied with the binary image to get the final segmentation result for computing the FgPA (see last column in Figure \ref{fig:segmentation_results}).

A further post-processing step to improve the FgPA computes all connected components in the result which are mostly single letters, and assigns the mode that is the most frequent label of all pixels in a component.
Therefore, the final outcome is a segmentation that predicts a consistent label for each connected component.
Obviously, regions of different types that are connected, e. g. image and text blocks, will introduce errors because only one label will be assigned to the complete component by this post-processing method.

\section{Experiments}

The performed experiments are conducted on all six available books.
To compute average measures each experiment is repeated ten times with randomly initialized network weights and randomly divided test and train sets.
The hyperparameters of the network structure and the solver are kept fixed during all experiments and across all books, which is why we do not optimize the method for each single book.
The books St. Gall, G. Washington, and Parzival provide a fixed testing and training split, which is exclusively used to compute the accuracy for comparison with other existing models.

The FCN is implemented in Caffe \cite{Jia:2014:Caffe} and runs on a Nvidia Titan X GPU.
A single image takes about 0.5s to process, which clearly improves the run time of the simple CNN approach by a factor of 6 to 10 compared to \cite{chen2017/convolutional}.
The optional post-processing runs on a i7-5820K 3.30GHz CPU and takes about 1-3s depending on the original image size.

\subsection{Metrics}
The metrics for evaluating the quality of a page segmentation are most commonly pixel-leveled, e. g. precision, recall or accuracy.
Other measures taking into account the area of a class are, e. g. the mean intersection over union (IU) or the frequency weighted IU (f.w. IU).
But even those more sophisticated measures yield different results for the three segmentations in Figure \ref{fig:equal_segmentations}.
Our novel metric overcomes this problem by taking only foreground pixels of the original binarized image into account.
Let $\delta_x = 1$ if the ground truth label at position $x$ matches the predicted label, else set $\delta = 0$.
Furthermore, define $b_x=1$ if $x$ is a foreground pixel else set $b_x = 0$.
Note that $\sum_x 1$ is the total number of pixels and $\sum_x b_x$ is the total number of foreground pixels.
We then compute the
\begin{itemize}
    \item Total Pixel Accuracy: $\text{TPA} = \frac{\sum_x \delta_x}{\sum_x 1}$
    \item Foreground Pixel Accuracy: $\text{FgPA} = \frac{\sum_x b_x \cdot \delta_x}{\sum_x b_x}$
    \item Foreground Pixel Error: $\text{FgPE} = 1 - \text{FgPA}$
\end{itemize}

\subsection{Evaluation}
All predicted segmentation images are rescaled to the full resolution of the original image before computing the pixel valued metrics.
The last column Figure \ref{fig:segmentation_results} shows two outcomes of the FCN after applying the bit mask (second column) for the G. Washington and GW5060 data sets as examples for the worst and best books, respectively.

\begin{table*}[!t]
    \caption{Total pixel accuracy TPA and foreground pixel accuracy FgPA in percent on different Methods where available. We use a 10-fold Monte Carlo cross-validation to obtain the results.}
    \label{tab:performance}
    \centering
    \begin{tabular}{c|cccccc}
    \hline
    \hline
        Data set & \multicolumn{2}{c}{G. Washington} & \multicolumn{2}{c}{Parzival} & \multicolumn{2}{c}{St. Gall} \\
                & TPA & FgPA & TPA & FgPA & TPA & FgPA \\
    \hline
        Local MLP \cite{conf/das/chen/2016} & $87$ & - & $91$ & - & $95$ & - \\
        CRF \cite{conf/icfhr/0011SLHLI16} & $91$ & - & $93$ & - & $97$ & - \\
        CNN \cite{chen2017convolutional} & $91$ & - & $\bm{94}$ & - & $\bm{98} $ & - \\
        FCN & $\bm{92.1}$ & $93.53 \pm 0.66$ & $\bm{93.6}$ & $96.75 \pm 0.47$ & $\bm{98.4}$ & $96.39 \pm 0.35$ \\
        FCN (Post-processed) & - & $93.7 \pm 1.1$ & - & $97.33 \pm 0.49$ & - & $97.67 \pm 0.39$ \\
    \hline
    \hline
        Data set & \multicolumn{2}{c}{GW5060} & \multicolumn{2}{c}{GW5064} &     \multicolumn{2}{c}{GW5066} \\
                & TPA & FgPA & TPA & FgPA & TPA & FgPA \\
    \hline
        FCN & $95.54 \pm 0.19$ & $98.64 \pm 0.12$ & $95.28 \pm 0.30$ & $96.20 \pm 0.70$ & $95.55 \pm 0.92$ & $94.71 \pm 0.61$ \\
        FCN (Post-processed) & - & $98.99 \pm 0.13$ & - & $96.48 \pm 0.69$ & - & $95.54 \pm 0.48$ \\
    \hline
    \end{tabular}
\end{table*}

\subsubsection{Accuracy of the FCN}
The established total pixel accuracy (TPA) of our FCN on the fixed splits of Parzival, St. Gall, and G. Washington are reported in Table \ref{tab:performance}.
On Parzival and St. Gall the FCN yields comparable results to a simple CNN approach \cite{chen2017convolutional} that is trained to predict the label of superpixels.
The FCN outperforms teh methods reported in \cite{conf/das/chen/2016} and \cite{conf/icfhr/0011SLHLI16} on the three examined data sets and the CNN approach on G. Washington.
On GW5060, GW5064 and GW5066 Table \ref{tab:performance} reports the TPA with its standard deviation on ten folds using a Monte Carlo cross-validation approach.

\subsubsection{Foreground Pixel Error}
The Foreground Pixel Error (FgPE) and its standard deviation computed on ten folds is listed in Table \ref{tab:performance}.
Although the total accuracy of the Parzival data set of 93.6\% is notably worse than the accuracy on St. Gall (98.4\%), the FgPA is comparable.
Therefore, the measure states that the segmentation result used for OCR is of similar quality.
A reason for this deviation in this example is found in the layout of the two books and the ground truth.
St Gall is written in justified text and its text masks are very similar to a simple rectangle.
In contrast Parzival's text is left aligned with varying line lengths.
Therefore, its masks that follow these indentations have a longer contour than a simple box.
Since most of the errors are located at the margin of a region, the longer contour is expected to produce more errors.
The FgPA ignores these margins and instead respects only foreground pixels which is why these different layouts do not affect this metric.

Similarly, the TPA on GW5060, GW5064, and GW5066 is comparable but the FgPA differs significantly, which is also explained by the different layouts of the books.
The TPA is mostly affected by the baseline, that is the amount of background pixels in a page.
This baseline is almost the same for GW5060 and GW5066.
GW5064 has a higher baseline but compared to the other two books its layout is the most difficult layout.
Text at the margin can be classified as either marginalia, running text or headline, depending on the context.
This neglects the effect of the higher baseline.

The FgPA is independent of the number of background pixels, which is why the results of the books are different.
The baseline of this metrics is the most frequent label of the foreground.
GW5066 has by far the lowest FgPA baseline and therefore reasonably has the worst FgPA score.
Both GW5060 and GW5064 have a similar baseline but since GW5064 has a difficult layout its score is lower than the one of GW5060.

\subsubsection{Training set size}
\begin{figure}[t]
\centering
\includegraphics[width=\linewidth]{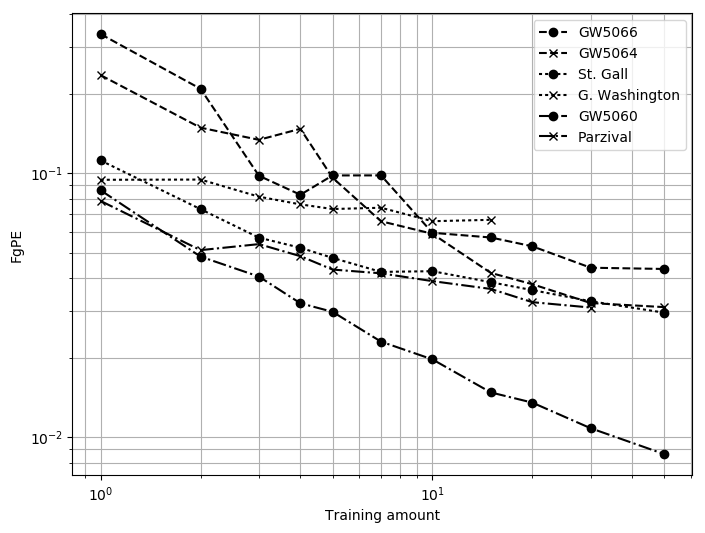}
\caption{The Foreground Pixel Error of the FCN on different absolute training set sizes. Note that G. Washington and Parzival only consist of 20 and 37 instances, respectively, which is why  data points are missing.}
\label{fig:fpge_abs}
\end{figure}

The number of pages in the training set that are required for a reliable segmentation result is a crucial quantity for the application of an algorithm because these are the pages that must be manually segmented.
We compute the performance of the FCN under different training set sizes by using a specific amount of all available pages in the training set and the rest in the test set.
Hereby we simulate a real world application scenario where a subset of a book is manually segmented and the resulting ground truth is used to train a model in order to process the remainder of the pages.
In the first experiment the number of pages $N$ in the training is chosen as absolute value $N \in \{1, 2, 3, 4, 5, 7, 10, 15, 20, 30, 50\}$, in a second run we chose $N$ relatively to the complete data set size $D$ with $N=r \cdot D$ and $r \in [0.05, 0.80]$ with step size $0.05$.
Figure \ref{fig:fpge_abs} clearly shows that the accuracy of all data sets is improved with an increasing amount of training data.

\begin{figure*}[!t]
\centering
\includegraphics[width=0.8\linewidth]{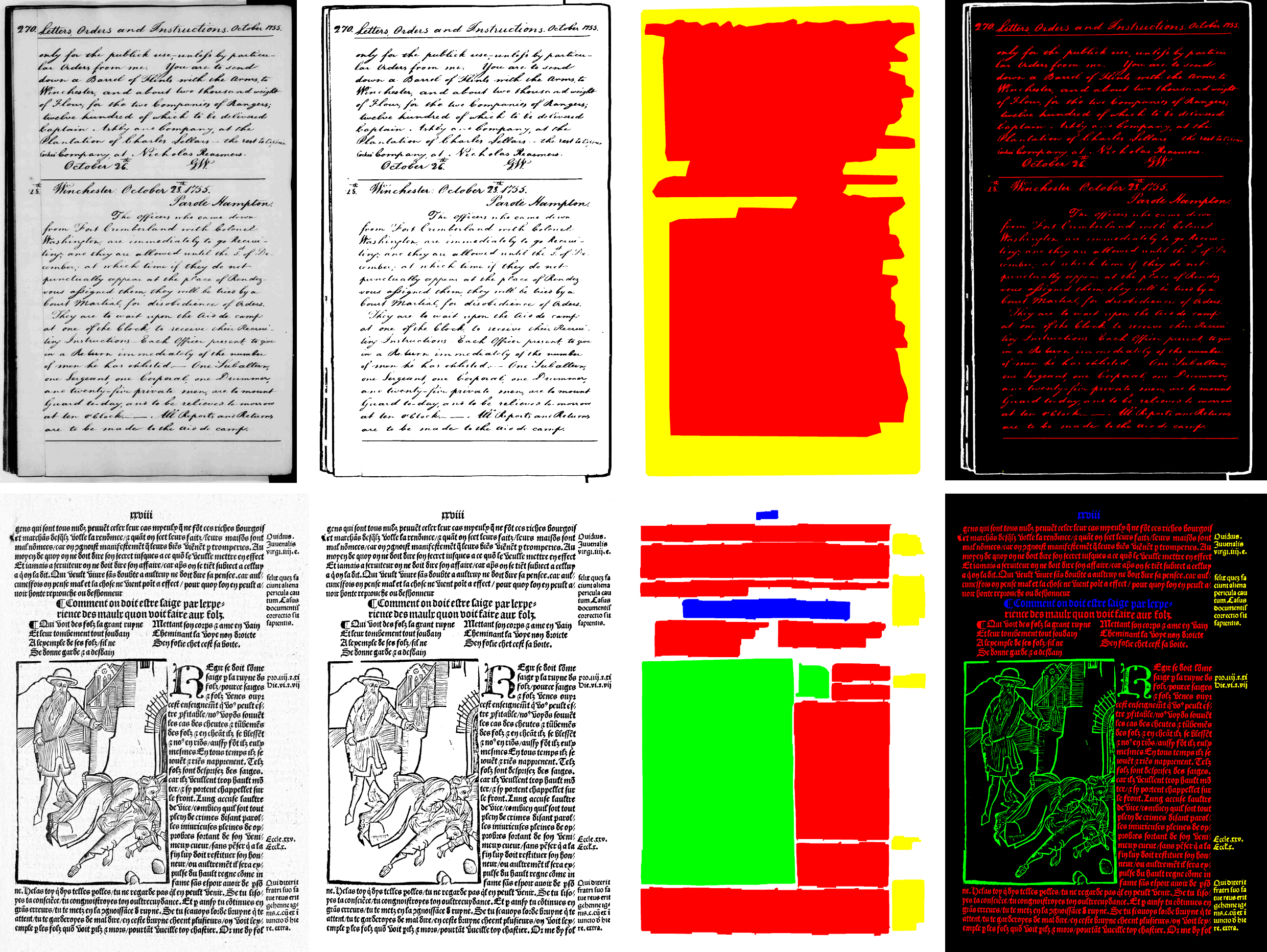}
\caption{Different images of the model. The columns from left to right show: the original grayscale image, the binarized image, the provided ground truth, the segmentation result. The first row is page 290 of the G. Washington data set using black as the ignored background, white as the periphery, yellow as page, and red as text. The second row shows the results page 55 of GW5060, using black as the ignored background, red as text, yellow as marginalia, blue as headlines, and green as images (best seen in color).}
\label{fig:segmentation_results}
\end{figure*}

The slopes of GW5060, GW5064 and GW5066 are in general steeper than those of G. Washington, Parzival, or St. Gall, what might be related to the different data set sizes or the different layouts.
The G. Washington data set has the smallest relative error decrease with increasing training examples.
The reason is that the ground truth of this data set in comparison with the others contains more errors, as can be seen in the first row of Figure \ref{fig:segmentation_results}:
The upper two horizontal lines in the ground truth are labeled as text, whereas the bottom line is labeled as page.
The learning algorithm can not learn if a detected line shall be classified as text or page.

\begin{figure}[t]
\centering
\includegraphics[width=\linewidth]{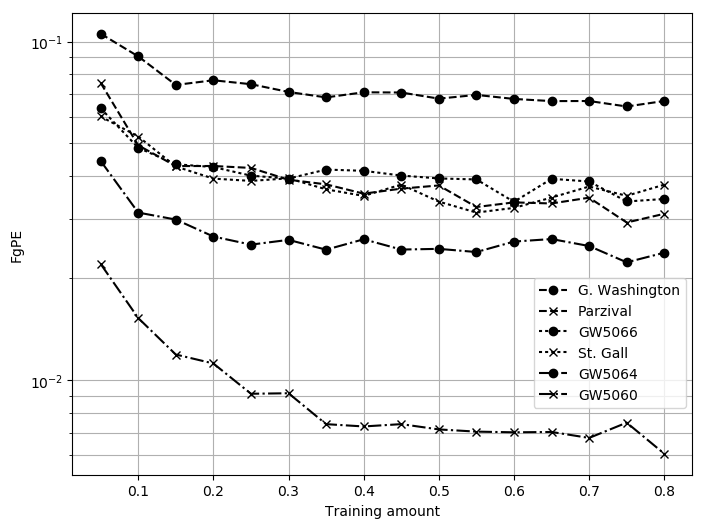}
\caption{The Foreground Pixel Error of the FCN on different relative training set sizes.}
\label{fig:fpge_rel}
\end{figure}

Figure \ref{fig:fpge_rel} shows similar results as Figure \ref{fig:fpge_abs}, but due to the huge differences in the number of total images in the data sets this chart shows the progression to very high amounts of training examples.
Since most of the curves in the log plot are almost flat starting at 20\%, additional pages in the training set do not improve the segmentation result.
Solely the GW5060 data set profits of a higher amount of training examples of up to 40\%, which interestingly is also the data set with the by far lowest FgPE.

\subsubsection{Manual selection of training images}

\begin{figure}[t]
\centering
\includegraphics[width=\linewidth]{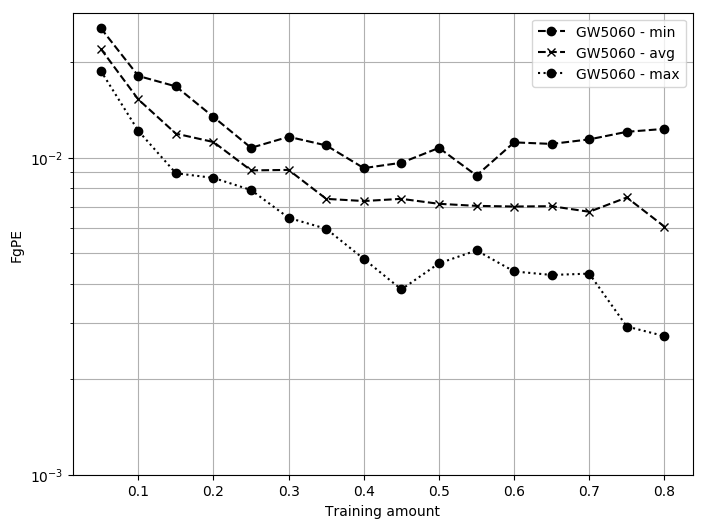}
\caption{The minimum, average and maximum FgPE for ten folds of the GW5060 data set on different relative training set sizes.}
\label{fig:fpge_rel_min_avg_max}
\end{figure}

If a user wants to train a model for page segmentation he has the freedom to choose which pages shall be used for training.
To estimate the effect of a useful and a poor selection we show in Figure \ref{fig:fpge_rel_min_avg_max} the minimum, average and maximum FgPE on the GW5060 data set for ten folds.
For only a few images in the training set, the difference between maximum and minimum  FgPE is clearly smaller than for a larger training corpus.
The reason is that at the small data set size any ordinary page is useful for training, whereas later in a larger data set special pages, e. g. the title page, should be segmented by hand.
Obviously in practice, pages that have a unique layout or contain unique marks or style must be segmented by hand.
Excluding these pages from the testing and even training data set will improve the result clearly towards the maximum curve in Figure \ref{fig:fpge_rel_min_avg_max}.

\subsubsection{Post-processing}
\begin{figure}[t]
\centering
\includegraphics[width=\linewidth]{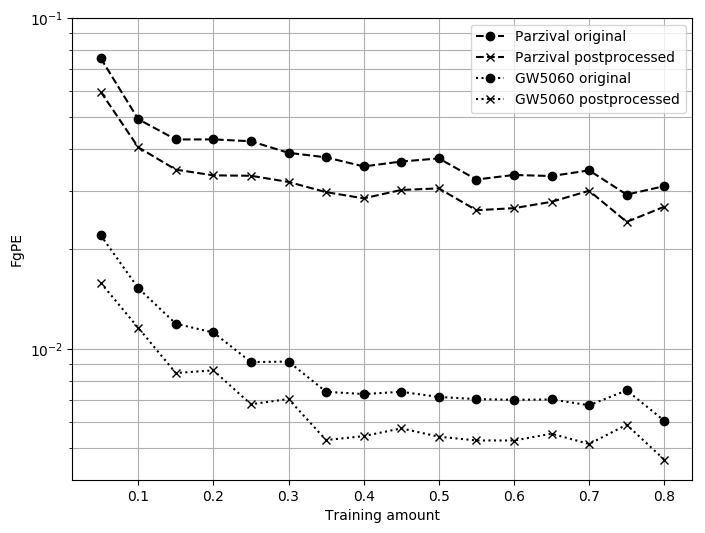}
\caption{The Foreground Pixel Error of the FCN on different relative training set sizes including post-processing. Only the results for GW5060 and Parzival are shown.}
\label{fig:fpge_rel_comp_post}
\end{figure}

Figure \ref{fig:fpge_rel_comp_post} shows the effect of the post-processing approach that chooses the most frequent label for a single connected component.
Obviously, the post-processing improves the results for different splits.
Although, the post-processing can join components that have different labels, e. g. an image and a character, the overall effect is positive
There the proposed post processing is meaningful.

\section{Conclusion}

We proposed a FCN for historical document segmentation.
This network learns and predicts a complete page in one step and does not require sophisticated preprocessing steps such as superpixels.
We achieved comparable or improved results on open source data sets and reduce the computation time by a factor of up to 10.

Furthermore, we introduced a novel meaningful metric to compare different models and methods of document segmentation by using the fact that only foreground information (black ink) is relevant for further processing steps in the pipeline of digitization of documents.

\IEEEtriggercmd{\enlargethispage{-0.5cm}}
\IEEEtriggeratref{12}
\bibliographystyle{IEEEtran}
\bibliography{literature}

\end{document}